\documentclass{article} %
\usepackage{iclr2026_conference,times}

\usepackage{amsmath,amsfonts,bm}

\def\eqref#1{equation~\ref{#1}}

\def\1{\bm{1}}

\DeclareMathAlphabet{\mathsfit}{\encodingdefault}{\sfdefault}{m}{sl}
\SetMathAlphabet{\mathsfit}{bold}{\encodingdefault}{\sfdefault}{bx}{n}

\newcommand{\websites}{\mathcal{W}}        %
\newcommand{\tasks}{\mathcal{T}}           %
\newcommand{\agentbrowser}{\mathcal{B}_{\text{browser}}} %
\newcommand{\agenttool}{\mathcal{B}_{\text{tool}}}       %

\newcommand{\actionsprim}{\mathcal{A}_{\text{prim}}}      %
\newcommand{\actionstools}{\mathcal{A}_{\text{tools}}}    %

\newcommand{\actclick}{a_{\texttt{click}}}
\newcommand{\acttype}{a_{\texttt{type}}}
\newcommand{\actnavigate}{a_{\texttt{navigate}}}

\newcommand{\tool}{u}                          %
\newcommand{\toolopt}{u^*}                     %
\newcommand{\toolcandidates}{\tilde{\mathcal{U}}} %
\newcommand{\toolcandidate}{\tilde{u}}         %

\newcommand{\traces}{\mathcal{X}}               %
\newcommand{\testinputs}{\mathcal{I}_{\text{test}}} %
\newcommand{\inputschema}{\mathcal{S}}          %
\newcommand{\schema}{\mathcal{S}_{\text{inp}}}  %
\newcommand{\feedback}{\mathcal{F}}             %

\newcommand{\failurerate}{\text{FailRate}}
\newcommand{\stepcount}{\text{StepCount}}
\newcommand{\agenticratio}{\text{AgenticRatio}}
\newcommand{\maxattempts}{N_{\text{max}}}
\newcommand{\goal}{\text{Goal}} %

\newcommand{\StabilizeSelectors}{\textsc{StabilizeSelectors}}   %
\newcommand{\AddAgenticFallbacks}{\textsc{AddAgenticFallbacks}} %
\newcommand{\ReplaceWithURLOps}{\textsc{ReplaceWithURLOps}}     %
\newcommand{\InferSchema}{\textsc{InferSchema}}                 %
\newcommand{\RefineCandidate}{\textsc{RefineCandidate}}         %

\newcommand{\methodname}{WALT} %

\usepackage{hyperref}
\usepackage{url}
\usepackage{xcolor}
\usepackage{colortbl}
\usepackage{amssymb}
\usepackage{graphicx}
\usepackage{booktabs}
\usepackage{multirow}
\usepackage{subcaption}
\usepackage{float}
\usepackage{caption}
\usepackage{algorithm}
\usepackage{algpseudocode}
\definecolor{ForestGreen}{rgb}{0.13, 0.55, 0.13}
\algdef{SE}[TRY]{Try}{EndTry}[1]{\textbf{try}}{\algorithmicend\ \textbf{try}}
\algdef{C}[TRY]{TRY}{Catch}[1]{\textbf{catch}}
\algtext*{EndTry}

\usepackage[most]{tcolorbox}
\usepackage{listings}
\usepackage{lipsum}
\usepackage{courier}
\title{\methodname{}: Web Agents that Learn Tools}

\author{%
\textbf{Viraj Prabhu} \quad
\textbf{Yutong Dai} \quad
\textbf{Matthew Fernandez} \quad
\textbf{Jing Gu} \quad
\textbf{Krithika Ramakrishnan} \\
\textbf{Yanqi Luo} \quad
\textbf{Silvio Savarese} \quad
\textbf{Caiming Xiong} \quad
\textbf{Junnan Li} \quad 
\textbf{Zeyuan Chen} \quad
\textbf{Ran Xu}\\
Salesforce AI Research
}

\iclrfinalcopy %
\begin{document}

\maketitle
\begin{abstract}
    Web agents promise to automate complex browser tasks, but current methods remain brittle—relying on step-by-step UI interactions and heavy LLM reasoning that break under dynamic layouts and long horizons. Humans, by contrast, exploit website-provided functionality through high-level operations like search, filter, and sort. We introduce \methodname{} (Web Agents that Learn Tools), a framework that reverse-engineers latent website functionality into reusable invocable tools. Rather than hypothesizing ad-hoc skills, \methodname{} exposes robust implementations of automations already designed into websites—spanning discovery (search, filter, sort), communication (post, comment, upvote), and content management (create, edit, delete). Tools abstract away low-level execution: instead of reasoning about \emph{how} to click and type, agents simply call \texttt{search(query)} or \texttt{create(listing)}. This shifts the computational burden from fragile step-by-step reasoning to reliable tool invocation. On VisualWebArena and WebArena, \methodname{} achieves higher success with fewer steps and less LLM-dependent reasoning, establishing a robust and generalizable paradigm for browser automation.
\end{abstract}

\section{Introduction}
\label{sec:intro}

\begin{figure}[htbp]
    \centering
    \includegraphics[width=\textwidth]{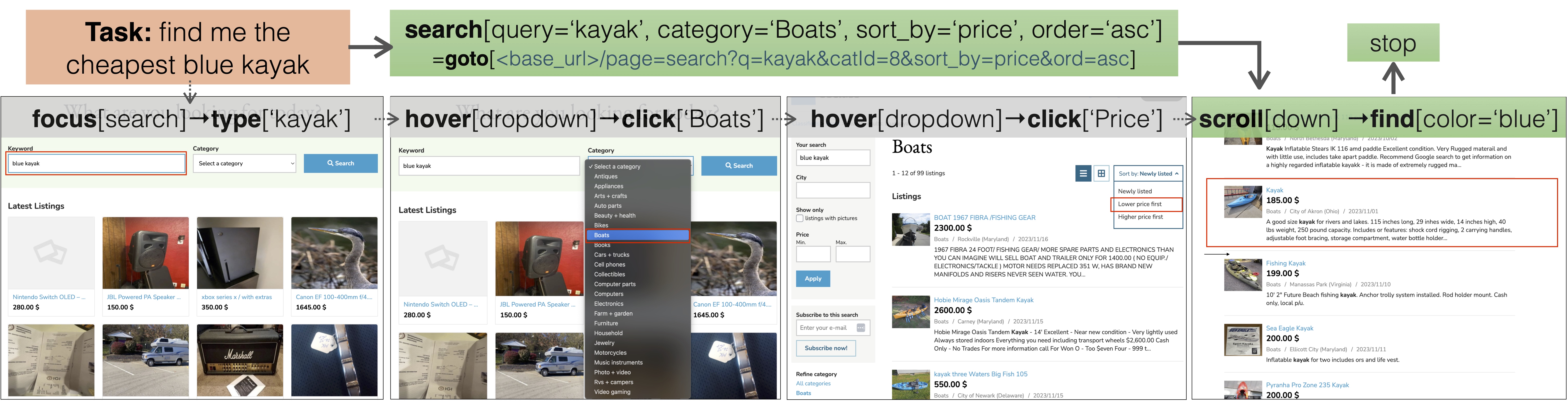}
    \caption{\textbf{\methodname{} transforms browser agent automation from brittle step-by-step reasoning to efficient tool-based abstraction.} Given the task ``find the cheapest blue kayak,'' traditional web agents execute a lengthy sequence of primitive UI actions focusing on search boxes, hovering over dropdowns, clicking categories, and sorting and scanning results. In contrast, our method \methodname{} (Web Agents that Learn Tools), designs a deterministic tool that exposes this website-provided functionality to the agent: \texttt{search(query=`blue kayak', category=`Boats', sort\_by=`price')}, reducing execution from 8+ fragile UI steps to 1 robust operation.}
    \label{fig:intro}
\end{figure}

Consider searching for the cheapest blue kayak on a classifieds page (Fig.~\ref{fig:intro}): existing web agents reason through each step—how to interact with the search box, locate filter controls, determine the correct sort option—while simultaneously handling implementation details like element selection and timing. In contrast, humans naturally think about this task in terms of website functionality: ``search for kayaks, filter by price, identify the first blue one.'' They abstract away the implementation details and focus on what they want to accomplish, not how the interface mechanics work.

This human capability stems from recognizing reusable patterns across websites — not only in search and filtering, but also in content creation and management (e.g., creating, editing, deleting listings) and social interactions (e.g., commenting, messaging, upvoting). Humans leverage this prior knowledge to quickly adapt their interaction strategies to new websites. In the web agent context, this intuition has inspired work on discovering ``skills''~\citep{wangagent} for web agents—reusable action sequences that encapsulate common interaction patterns and can be applied across similar website elements or tasks.

However, existing skill discovery approaches suffer from two key limitations in \emph{which} skills are discovered and \emph{how} they are implemented. First, they either mine skills only from successful trajectories~\citep{wangagent,sarch2024vlm,wang2025inducing}—codifying existing behaviors—or require agents to hypothesize useful automations~\citep{zheng2025skillweaver}, often yielding unintuitive, overly specific, or irrelevant skills. Second, both approaches implement skills as brittle UI action sequences, highly sensitive to dynamic elements and design changes. 

We propose \methodname{} (Web Agents that Learn Tools). Unlike prior “skills” or “workflows,” which are agent-induced action sequences, our tools correspond to \emph{website-provided functionality}—search bars, filters, sorting mechanisms, commenting systems, and navigation controls—that site designers have already engineered as robust automations. Each tool is exposed to the agent as a high-level deterministic call, with an underlying implementation discovered and validated through reverse-engineering. This reframing shifts the agent's capability frontier: instead of learning brittle approximations of interaction patterns, \methodname{} surfaces the functionality already embedded in websites as reliable, reusable tools.

On each website, \methodname{} follows a demonstrate-generate-validate loop for each identified tool: (1) a web agent comprehensively demonstrates the functionality (\emph{e.g.}, all filters and sort options for search); (2) a tool generation agent maps execution traces to structured tools with validated input schemas, prioritizing deterministic actions but allowing agentic steps for dynamic elements, and attempting to replace UI sequences with more robust URL manipulation through API reverse-engineering; (3) a test agent verifies functionality against pre-vetted test inputs. 

This abstraction transforms the agent's computational burden: instead of reasoning about ``how do I search for X, then filter by Y, then sort by Z'' through complex UI sequences, the agent simply calls \texttt{search(X)}, \texttt{filter(Y)}, \texttt{sort(Z)} and focuses on higher-level planning. Tool discovery and optimization happen offline during website exploration, ensuring both efficiency and reliability.

We benchmark our method on VisualWebArena~\citep{koh2024visualwebarena} and WebArena~\citep{zhouwebarena}, discovering over 50 reusable tools spanning search and filtering, content creation and management (e.g., create, edit, delete listings), and communication or social interactions (e.g., commenting, messaging, upvoting). \methodname{} achieves state-of-art success rates of 52.9\% on VisualWebArena and 50.1\% on WebArena, significantly outperforming prior work. Ablation studies further reveal that our proposed contributions — discovered tools, multimodal DOM parsing, and external verification -- yield gains in both success rates (10\%-30\% across splits) and efficiency (1.3-1.4x fewer steps on average). Overall, \methodname{} transforms browser agent automation from brittle step-by-step reasoning to efficient tool-based abstraction.

\section{Related Work}
\label{sec:relwork}

\textbf{Web Agents.} Agents capable of directly operating a browser to perform tasks hold promise for automating online tasks. Prior work advances web agents along four axes: \textbf{Perception} concerns what the agent sees and how it grounds elements: some methods parse raw HTML~\citep{gur2022understanding,deng2023mind2web}, others process full-page screenshots with vision-language models~\citep{furuta2023multimodal,he2024webvoyager}, often augmented by Set-of-Mark (SoM) visual prompts~\citep{yang2023set}; recent work improves page understanding and grounding via prompting~\citep{zheng2024gpt,yao2023react} and task-specific training~\citep{furuta2023multimodal,zhang2025tongui,pahuja2025explorer,qi2024webrl}. \textbf{Planning} scales test-time exploration with search (e.g., MCTS and related variants) to choose better action sequences~\citep{koh2024tree,putta2024agent,yu2410teaching,gu2024your}. \textbf{Reasoning} enhances step selection through chain-of-thought and ReAct-style prompting~\citep{wei2022chain,yao2023react}. \textbf{Action execution} determines how decisions touch the page: agents that use HTML or SoM predict DOM targets, whereas screenshot-only agents act via pixel-space coordinates~\citep{xu2024aguvis}, which can be more brittle to layout change. Our approach targets the action-execution and planning axes by mining reusable, efficient \emph{tools} offline—encapsulating site functionality with validated schemas, URL-level operations, and targeted agentic fallbacks—so agents solve tasks faster and more reliably than step-by-step UI policies.

\textbf{Benchmarks for Web Agents.} Benchmarks for web agents are expanding rapidly and span simulated and real environments. Early simulated testbeds~\citep{shi2017world, liu2018reinforcement} emphasize basic navigation such as clicking and form filling, whereas \citet{yao2022webshop} focused on e-commerce tasks. ~\cite{zhouwebarena} introduced WebArena, a realistic web simulation environment with replicas of various website types (\emph{e.g.} shopping, public forum, maps, etc.), with rich functionality and realistic underlying databases, and a set of complex natural language tasks paired with robust rule-based evaluators. VisualWebArena~\citep{koh2024visualwebarena} further extended this environment to include visually-grounded tasks that require rich multimodal understanding, along with supporting website and evaluator additions. A complementary direction evaluates agents on real websites or production sandboxes~\citep{he2024webvoyager,zhang2025functionality,drouin2024workarena,boisvert2024workarena++}, covering e-commerce, enterprise software, and everyday workflows. We focus on WebArena and VisualWebArena, and propose a method to autonomously discover and construct reusable, website-specific tools that significantly improve agent performance and efficiency.

\textbf{API-using Web Agents.} While UI-level actions are the default interface to the Web, they can be inefficient and brittle. Accordingly, some works exploit API documentation to design high-level actions from APIs and thus augment or bypass UI interactions~\citep{song2024beyond,ni2025doc2agent}. In contrast, we do not assume any API documentation -- which is often undocumented or proprietary -- and instead attempt to reverse-engineer website-provided functionality into callable tools with validated input schemas, URL-parameter promotion, and agentic recovery, all learned autonomously via systematic exploration.

\textbf{Skill Discovery for Web Agents.} Some recent works focus on discovering skills for web agents by mining successful agent trajectories: SkillWeaver~\citep{zheng2025skillweaver} produces unit-tested Python functions from successful attempts, whereas AWM~\cite{wangagent} and ASI~\citep{wang2025inducing} induce skills online (represented as text and programs, respectively) by prompting an agent to induce skills from action subsequences in successful trajectories. Both lines of work typically mine only from successful executions and implement skills by composing primitive actions, which can be brittle and effectively codify current behavior without expanding capability. By contrast, we systematically explore website-specific functionality and exploit observable regularities and site infrastructure; our learned tools are stress-tested and iteratively optimized for reliability and modularity. Unlike prior work that composes longer UI sequences, we discover and implement new, website-grounded tools with schema validation, selector stabilization, URL reverse-engineering, and targeted agentic fallbacks.

\section{Approach}
\label{sec:approach}

\begin{figure}[t]
    \centering
    \includegraphics[width=\textwidth]{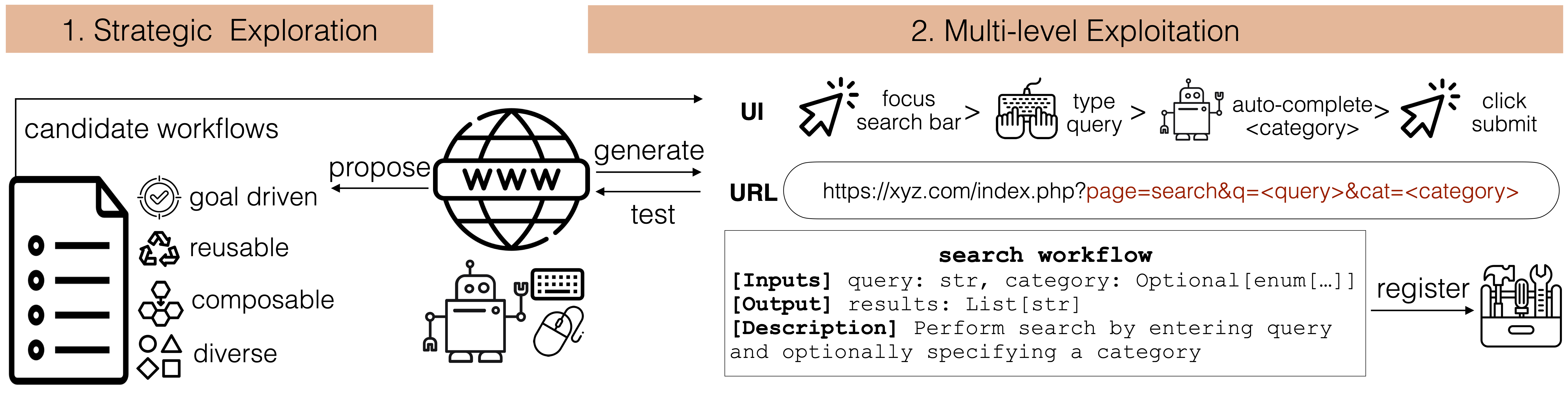}
    \caption{Overview of WALT. \textbf{Left—Discovery:} the browser agent explores key site sections to propose tool candidates and record stabilized interaction traces (robust selectors with fallbacks). \textbf{Right—Construction \& validation:} the tool constructor turns traces into an action script (navigation, extraction, interaction, agentic steps), promotes eligible UI chains to URL operations, induces a validated input schema, then registers and tests the tool end to end. Feedback refines selectors, schema, and script until a robust single-call tool is produced.}
    \label{fig:approach}
\end{figure}

We frame browser automation as the discovery and use of \emph{tools}—high-level, callable operations that abstract away fragile low-level interactions. Unlike prior work that induces ad-hoc skills or scripted action sequences, \methodname{} treats websites as sources of structured functionality (e.g., search, filter, post). Each tool is backed by a validated action script -- primarily deterministic URL/DOM operations with targeted agentic steps - Figure~\ref{fig:approach} summarizes the two-stage pipeline: strategic discovery of tool candidates followed by their construction and validation.

\subsection{Problem Formulation}

Let $\websites = \{w_1, w_2, \ldots, w_n\}$ denote a set of websites, and $\tasks = \{t_1, t_2, \ldots, t_m\}$ denote a set of tasks. A \textbf{browser agent} $\agentbrowser$ typically solves these tasks using primitive actions $\actionsprim = \{\actclick, \acttype, \actnavigate, \ldots\}$. Our goal is to discover and implement \textbf{tools} that can be registered as high-level actions $\actionstools$, enabling $\agentbrowser$ to operate with an expanded action space $\actionsprim \cup \actionstools$ for more efficient and reliable task execution.

We define a \textbf{tool} $\tool$ as a callable high-level action $\tool : \inputschema \rightarrow \goal$ where $\inputschema$ specifies structured input parameters and $\goal$ is the target outcome. Once validated, tools are exposed to the policy as atomic actions that augment the agent's vocabulary.
Our approach involves two stages:

\textbf{Stage 1: Tool Discovery} systematically explores user-facing website sections to identify reusable functionality patterns. The process begins by prompting a web agent to navigate to key sections (content, discovery, communication areas) and discovering interactive elements through targeted interactions such as hovering over dropdowns and clicking menus. Finally, it is tasked with strategically proposing a list of reusable tool candidates with clear user intent that maximize coverage and minimize redundancy.

\textbf{Stage 2: Tool Construction \& Validation} transforms discovered tool candidates $\toolcandidate$ into validated, executable tools through a two-agent system. The process introduces a specialized \textbf{tool construction agent} $\agenttool$ that collaborates with the \textbf{browser agent} $\agentbrowser$: the browser agent executes candidates to produce traces, while the tool construction agent analyzes these traces and induces structured action scripts. Candidate tools are then validated against test inputs before being exposed as callable high-level actions.
\subsection{Algorithm Design}

\paragraph{Objective.}
We formulate tool construction as a multi-objective optimization problem. First, the tool discovery phase generates a set of $K$ tool candidates $\toolcandidates = \{\, (s_i, E_i, G_i) \mid i = 1,\ldots,K \,\}$ where $s_i$ is the start URL, $E_i$ are the relevant interactive elements, and $G_i$ is the specific goal to be accomplished. For each candidate, the browser agent $\agentbrowser$ first \emph{demonstrates} the functionality to produce a grounded execution trace $\traces$, which a separate tool construction agent $\agenttool$ then analyzes to synthesize an executable tool and associated test inputs. Formally,
\begin{align}
\text{given} \quad & \toolcandidate = (s_i, E_i, G_i) \quad\ \\
\text{execute \& generate} \quad & \toolcandidate \xrightarrow{\agentbrowser} \traces \xrightarrow{\agenttool} (\tool, \testinputs) \quad\ \\
\text{minimize} \quad & \failurerate(\tool,\testinputs) + \stepcount(\tool) + \agenticratio(\tool).
\end{align}
Here, $\failurerate$ is the fraction of failing test cases (measuring correctness), $\stepcount$ is the number of primitive operations the implementation executes (measuring efficiency), and $\agenticratio$ is the fraction of steps that require LLM-dependent reasoning (measuring determinism). The process iterates—updating the tool and test set with feedback—until a validated $\toolopt$ is obtained or the attempt budget is exhausted.

\paragraph{Tool Construction.} Each tool is realized as an \emph{action script}: a finite sequence of \emph{navigation}, \emph{extraction}, \emph{UI interaction}, and \emph{agentic} steps that the browser agent executes automatically. We seek to expose tools as atomic actions: the agent calls them by contract and relies on their internal execution, without having to reason about intermediate steps. We deliberately bias tool execution towards deterministic operations (navigation and interaction) to improve robustness and efficiency, but permit agentic steps when interfaces are dynamic or ambiguous. We demonstrate the generality of this pipeline across constructing a diverse range of tools for discovery (e.g., search, filter, sort), content management (create, edit, delete), and communication (comment, message, upvote). 

The tool construction pipeline unfolds in three tightly coupled phases:

\textbf{Tool demonstration} begins with $\agentbrowser$ executing the candidate to collect traces $\traces$ of primitive interactions under realistic timing, focus, and layout conditions. Rather than replaying brittle action sequences as in prior workflow/skill approaches~\citep{wangagent,sarch2024vlm,zheng2025skillweaver}, we explicitly prioritize discovering stable DOM selectors and store element hashes to allow for deterministic replay. This ensures subsequent execution targets elements that remain valid under minor UI changes. During exploration we also record viable alternative selectors, yielding vetted backups without overfitting to a single locator.

\textbf{Tool generation} maps execution traces $\traces$ into a stepwise \emph{action script} in two passes. In the first pass, four types of steps are supported: \textbf{UI interaction} uses cached element hashes from the execution trace for precise element targeting, with alternate selectors as backup. \textbf{Navigation} steps allow for changing URL routes or query parameters, which are then used to extract DOM contents for subsequent steps. \textbf{Extraction} steps allow for extracting DOM state for subsequent steps. \textbf{Agentic} steps are used for dynamic interactions that may be difficult to deterministically model, such as lazy-loaded content or file uploads. The multiple safeguards built into the action script ensure higher robustness and recoverability than prior work. 

In the second pass, $\agenttool$ attempts to further optimize the existing action script by reverse-engineering parameterizable URL routes when the site exposes them (\emph{e.g.} search query parameters). If successful, this replaces multi-step UI subsequences with a single URL manipulation, shortening execution and typically reducing both $\stepcount(\tool)$ and $\agenticratio(\tool)$ -- an optimization missing in earlier work. Finally, $\agenttool$ induces an input schema $\schema$ with rich validation and support for diverse datatypes (\emph{e.g.} enums for dropdown menus), optional fields, and usage examples, as well as a detailed tool description that specifies the tool's intended usage, preconditions, and expected outcomes.

\textbf{Tool registration and validation} closes the loop. We register $(\tool,\schema,\testinputs)$ as a callable action and execute it end-to-end with a new $\agentbrowser$ over pre-vetted $\testinputs$. Failures yield structured feedback $\feedback$ (selector drift, uncovered enum values, timing issues, or semantic mismatches after URL promotion) which $\agenttool$ uses to refine selectors (preferring the most stable hashes observed), amend $\schema$ (e.g., adding missing options or relaxing optionals), or edit the action script (e.g., relocating a selector or backing off an over-aggressive URL substitution). This closed validation loop systematically improves correctness and robustness, in contrast to one-shot script extraction. 

Only tools that pass validation within a fixed number of attempts are exposed to the agent at runtime. As an additional failsafe for the deployed agent against unanticipated failures (\emph{e.g.} significant UI changes), we equip the tool-enriched agent with \emph{agentic fallback} - a last resort that spawns a fresh agent to handle a failing action script on the fly. Finally, we equip the agent with two more generic tools -- a multimodal DOM parser that converts an HTML page to an interleaved input that simplifies cross-modal reasoning, and an external verification tool that we use to corroborate the agent's self-reported task outcome, following the verifier design proposed in \citet{xue2025illusion}.

In this manner, \methodname{} turns complex website functionality into simple tool calls. By pairing grounded interaction ($\agentbrowser$) with schema-checked, URL-optimized executors ($\agenttool$), it delivers robust tools across discovery, content, and communication that run faster and with fewer LLM calls.

\section{Experiments}
\label{sec:experiments}

\subsection{Overview}

We evaluate \methodname{} on two established web agent benchmarks: VisualWebArena~\citep{koh2024visualwebarena} and WebArena~\citep{zhouwebarena}. Our experiments demonstrate that \methodname{} achieves significant improvements over prior state-of-the-art methods by leveraging website-provided tools rather than brittle UI interaction sequences, improving success rates while reducing action steps. We conduct comprehensive ablation studies to validate the contribution of each component and provide detailed analysis of when and why \methodname{} succeeds.

\subsection{Benchmarks}

\textbf{VisualWebArena} contains 910 visually-grounded and human-annotated web-tasks instantiated in three highly-realistic and fully-featured websites -- Classifieds (234), Shopping (466), and Reddit (210). \textbf{WebArena} includes 812 more general tasks spanning five websites (two of which overlap with VisualWebArena) -- GitLab (180), Map (109), Shopping (187), CMS (also referred to as Shopping Admin - 182), Reddit (106), and Multi-site (48). Tasks are defined by a human-annotated intent (e.g. ``find the cheapest blue kayak and return its URL'') and evaluator functions (e.g. ``\texttt{assert URL == <XYZ>}''). Besides a robust set of (exact, inclusion, and fuzzy) string and URL matching, the benchmarks also support sophisticated evaluators based on parsing page HTML and image contents. Agents are evaluated by their binary success rate -- a stringent metric that only considers task completion rather than partial success, and is measured objectively by the evaluator function rather than a subjective LLM judgement.

\subsection{Implementation Details}

For our base agent, we pair a VLM planner (GPT-5~\cite{openai2025gpt5}) with a browser action executor (GPT-5-mini) using standard web actions (click, type, navigate, etc.). Observations include a page screenshot with indexed Set-of-Mark (SoM) boxes and a list of interactive elements keyed by the same indices. State is maintained via a multimodal message queue. For retrieval, we store trajectory summaries in a vector database keyed by task intent; at run time we embed the current intent and append the nearest summary in the DB (with similarity threshold $0.3$) as additional context. Agents authenticate to each site before execution, run for at most 30 steps, and replan every 15. We use GPT-5-mini as the verification LLM following the design of WebJudge~\citet{xue2025illusion}. The multimodal DOM parser converts a markdown dump of the page into an interleaved representation. Implementations build on browser-use~\citep{browseruse2024} and workflow-use libraries~\citep{workflowuse2024}.

\subsection{Baselines}

We compare against a representative set of state-of-the-art methods :

- \textbf{Skill-based web agents}: Specifically, on WebArena we benchmark against SkillWeaver~\citep{zheng2025skillweaver}, AWM~\citep{wangagent}, and ASI~\citep{wang2025inducing}. On VisualWebArena, we benchmark against concurrent world in tool-oriented web agents~\cite{yu2025aworld}.

- \textbf{Web agents with test-time scaling}: We benchmark against methods that use MCTS~\citep{koh2024tree} and reflective-MCTS~\citep{yu2410teaching}, as well as one that uses model-based planning~\citep{gu2024your}.

- \textbf{API-using web agents}: We benchmark against Hybrid Agent~\citep{song2024beyond}, which generates actions from API documentations curated for WebArena.

- \textbf{Computer-Use Agents}: Specifically, we benchmark the Claude Computer-Use Agent~\citep{claudecua}, implementation details in Appendix~\ref{subsec:implementation_appx}.

Additionally, we benchmark against SGV~\citep{andrade2025let}, which proposes using an external verification module to mitigate LLM agreement bias. Finally, we include strong baselines from the original benchmark papers as well as human performance as an upper bound.

\subsection{Main Results}

\begin{figure}[t]
    \centering
    \begin{minipage}[t]{0.6\textwidth}\vspace{0pt}
        \centering
        \resizebox{\textwidth}{!}{
        \begin{tabular}{@{}lcccc@{}}
        \toprule
        \textbf{Method} & \textbf{Classifieds} & \textbf{Shopping} & \textbf{Reddit} & \textbf{Avg.} \\
        \midrule
        GPT-4V+SoM~\citep{koh2024visualwebarena} & 9.8 & 17.1 & 19.3 & 16.4\\
        TreeSearch~\citep{koh2024tree} & 26.5  & 29.0 & 20.5 & 26.4\\
        WebDreamer~\citep{gu2024your} & 25.0& 26.3 & 15.9 & 23.2 \\
        Computer-Use~\citep{claudecua} & 36.7 & 21.9  & 27.5 & 27.0   \\
        ExaCT~\citep{yu2410teaching} & 41.0 &32.3 & 28.7 & 33.7\\
        AWorld~\cite{yu2025aworld} & -& - & - & 36.5 \\    
        SGV~\citep{andrade2025let} & 52.0 & \textbf{57.0}  & 33.0 & 50.2  \\
        \rowcolor{blue!5} 
        \methodname{} (Ours) & \textbf{64.1} & 53.4 & \textbf{39.0} & \textbf{52.9} \\
        \midrule
        \textcolor{gray}{Human}~\citep{koh2024visualwebarena} & \textcolor{gray}{91.7} & \textcolor{gray}{88.4} & \textcolor{gray}{87.1} & \textcolor{gray}{\textbf{88.7}} \\
        \bottomrule
        \end{tabular}
        }
    \end{minipage}
    \hfill
    \begin{minipage}[t]{0.38\textwidth}\vspace{0pt}
        \centering
        \includegraphics[width=\textwidth]{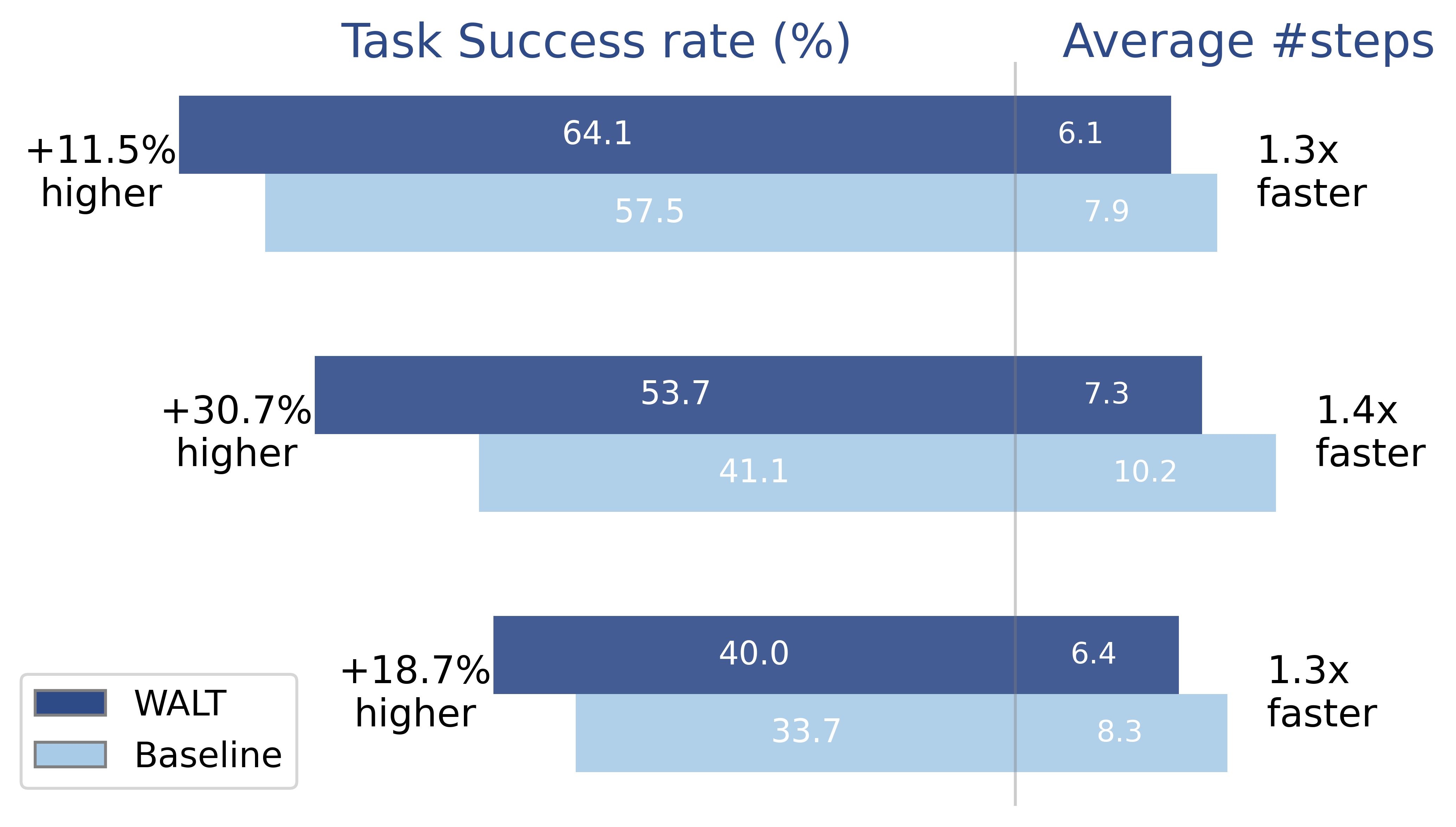}
    \end{minipage}
    \caption{\label{fig:results_vwa}\textbf{Results on VisualWebArena.} \textbf{Left.} We report success rate (\%) on each split as well as a weighted average. \textbf{Right.} We compare \methodname{}'s performance and efficiency with a baseline implementation as control.}    
\end{figure}

\begin{table}[t]
    \centering
    \resizebox{\textwidth}{!}{
    \begin{tabular}{@{}lcccccccc@{}}
    \toprule
    \textbf{Method} & \textbf{Gitlab} & \textbf{Map} & \textbf{Shopping} & \textbf{CMS} & \textbf{Reddit} & \textbf{Multi} & \textbf{Avg.} \\
    \midrule
    GPT-4+CoT~\citep{zhouwebarena} & - & - & - & - & - & - & 14.4 \\
    SkillWeaver~\citep{zheng2025skillweaver} & 22.2 & 33.9 &  27.2 &  25.8 &  50.0 & - &  \textcolor{gray}{29.8} \\
    AWM~\citep{wangagent} & 28.9 & 39.4 & 34.8  & 39.0 &  51.9 & 18.8 &35.5 \\    
    ASI~\citep{wang2025inducing} & 32.2 & 43.1 & 40.1 & 44.0& 54.7 & 20.8&40.4 \\
    Hybrid Agent~\citep{song2024beyond} & 44.4 & 45.9 & 25.7 & 41.2 & 51.9 & 16.7 & 38.9\\    
    \rowcolor{blue!5} 
    \rowcolor{blue!5} 
    \methodname{} (Ours) & 57.0 & 58.7 & 41.2 & 56.2 &48.5 & 20.8 & 50.1 \\
    \midrule
    \textcolor{gray}{Human}~\citep{zhouwebarena} & - & - & - & - & - & - & \textcolor{gray}{\textbf{78.2}} \\
    \bottomrule
    \end{tabular}
    }
    \caption{Performance comparison on WebArena benchmarks showing success rates (\%) across different domains. Bold values indicate best performance in each column.\label{tab:results_wa}}    
\end{table}

\begin{table}[t]
    
    \caption{Ablations on VisualWebArena-Classifieds showing the impact of different components on success rate (SR) and average number of steps. Results shown for different LLM backbones.\label{tab:ablations}}    
    \resizebox{\textwidth}{!}{
    \centering
    \begin{tabular}{@{}llccccl@{}}
    \toprule
    \textbf{browser LLM} & \textbf{tools} & \textbf{dom-parser} & \textbf{verify} & \textbf{avg \#steps ($\downarrow$)} & \textbf{SR (\%) $\uparrow$} \\
    \midrule    
    gpt-4.1 & none & text & self &  7.6 & 34.9   \\ %
    gpt-4.1 & discovered & text & self& \;\;\;\;\;\;\;\;\;\;6.6$_{\textcolor{ForestGreen}{-13.1\%}}$ & \;\;\;\;\;\;\;\;36.4$_{\textcolor{ForestGreen}{-4.3\%}}$ \\ %
    \specialrule{0.2pt}{0.5pt}{0.5pt}
    gemini-2.5-flash & none & text & self & 10.5 & 52.6 \\ %
    gemini-2.5-flash & discovered & text & self & \;\;\;\;\;\;\;\;\;\;\;8.3$_{\textcolor{ForestGreen}{-26.5\%}}$ & \;\;\;\;\;\;\;\;55.3$_{\textcolor{ForestGreen}{+5.1\%}}$ \\ %
    \specialrule{0.2pt}{0.5pt}{0.5pt}
    gpt-5-mini & none & text & self & 8.9 & 57.5 \\ %
    gpt-5-mini & discovered & text & self& \;\;\;\;\;\;\;\;\;\;\textbf{6.5}$_{\textcolor{ForestGreen}{-27.0\%}}$ & \;\;\;\;\;\;\;\;\;61.5$_{\textcolor{ForestGreen}{+7.0\%}}$ \\ %
    \textcolor{gray}{gpt-5-mini}& \textcolor{gray}{human demo} & \textcolor{gray}{text} & \textcolor{gray}{self}& \;\;\;\;\;\;\;\;\;\;\textcolor{gray}{7.4}$_{-16.9\%}$ & \textcolor{gray}{\;\;\;\;\;\;\;\;\;\;66.0}$_{+16.2\%}$ \\ %
    gpt-5-mini &  none & multimodal & self & \;\;\;\;\;\;\;\;\;\;7.5$_{\textcolor{ForestGreen}{-15.7\%}}$  & \;\;\;\;\;\;\;\;59.0$_{\textcolor{ForestGreen}{+2.6\%}}$ \\ %
    gpt-5-mini &  none &  text & external & \;\;\;\;\;\;\;\;\;11.0$_{\textcolor{red}{+23.6\%}}$ & \;\;\;\;\;\;\;\;59.4$_{\textcolor{ForestGreen}{+3.3\%}}$\\ %
    \rowcolor{blue!5}
    gpt-5-mini& discovered & multimodal & external &  \;\;\;\;\;\;\;\;\;\;7.0$_{\textcolor{ForestGreen}{-21.3\%}}$ & \;\;\;\;\;\;\;\;\;\;\textbf{64.1$_{\textcolor{ForestGreen}{+11.5\%}}$}\\ %
    \bottomrule
    \end{tabular}
    }
\end{table}

\begin{figure}[t]
    \centering
    \begin{subfigure}[b]{0.43\textwidth}
        \centering
        \begin{subfigure}[t]{\textwidth}
            \centering
            \includegraphics[width=\textwidth]{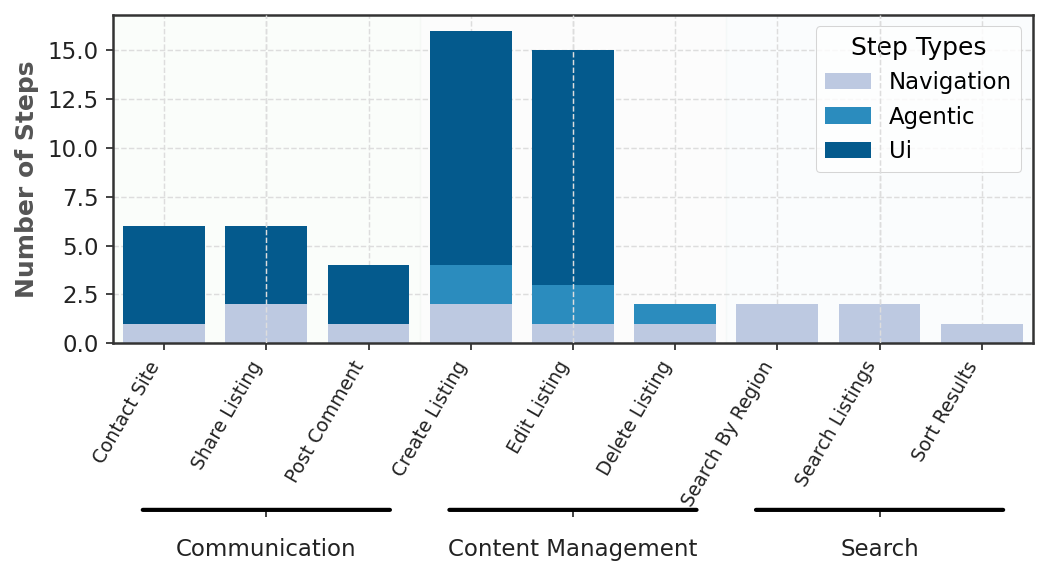}
            \caption{Task type and complexity breakdown}
            \label{fig:workflow_stats}
        \end{subfigure}
        \vspace{0.5em}
        \begin{subfigure}[b]{\textwidth}
            \centering
            \includegraphics[width=\textwidth]{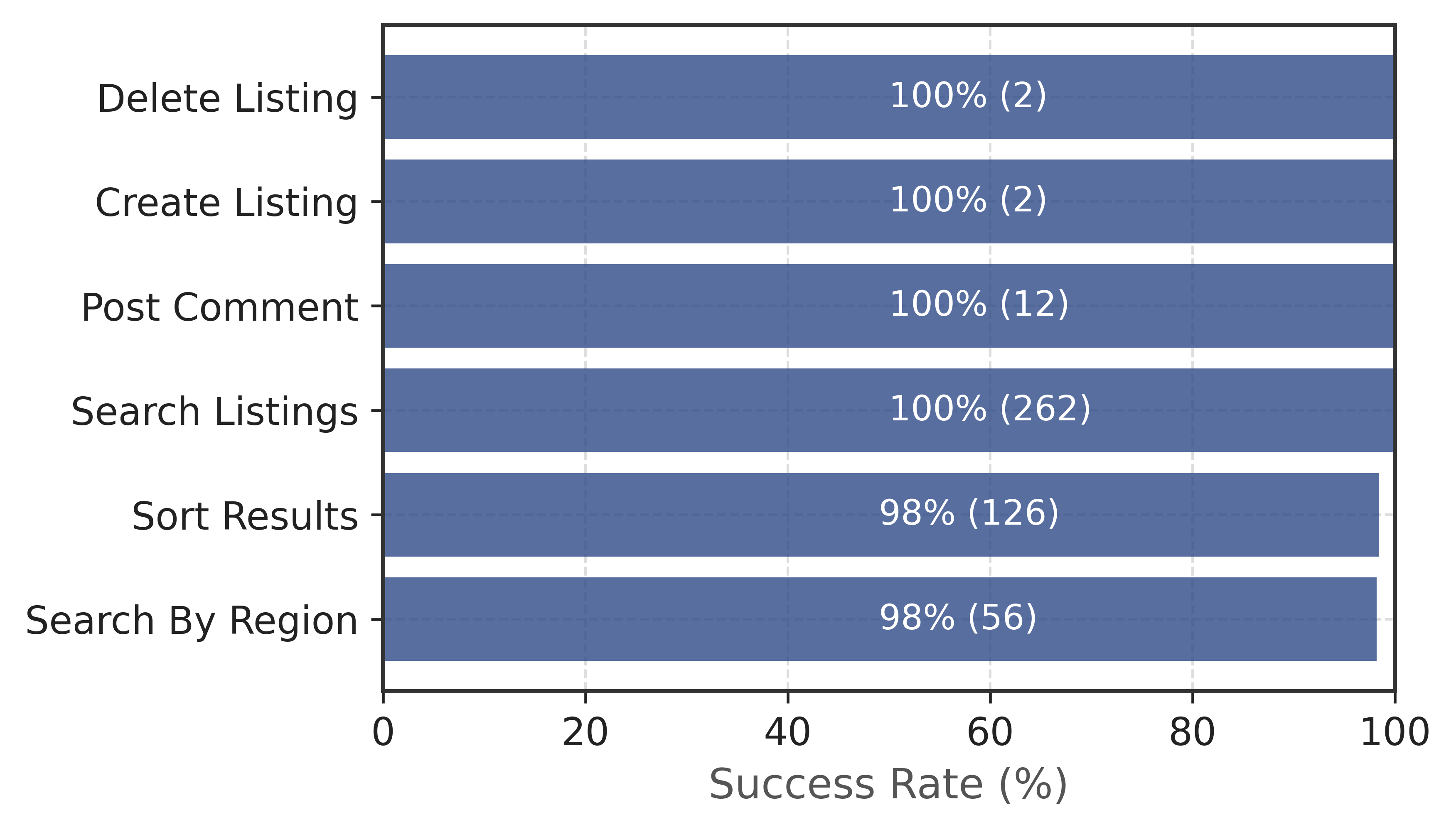}
            \caption{Success rate analysis by task complexity}
            \label{fig:success_rate_analysis}
        \end{subfigure}
    \end{subfigure}
    \hfill
    \begin{subfigure}[b]{0.54\textwidth}
        \centering
        \includegraphics[width=\textwidth]{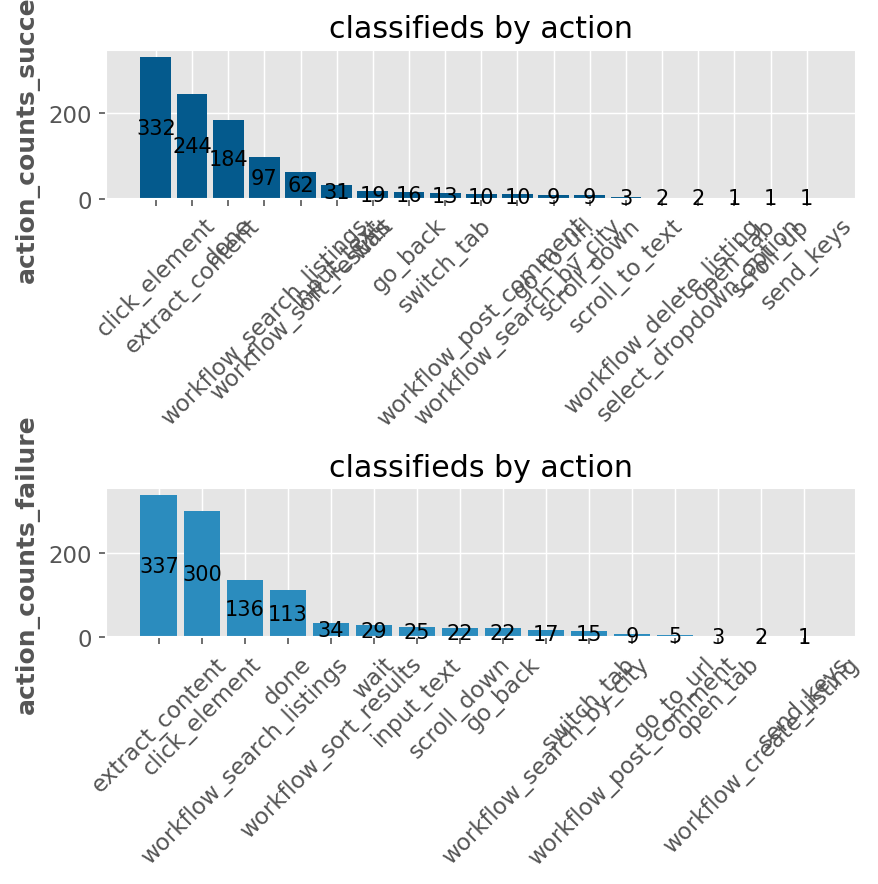}
        \caption{Action type analysis}
        \label{fig:classifieds_action}        
    \end{subfigure}

    \caption{Detailed analysis of the composition, success rates, and runtime invocations of tools discovered on the VisualWebArena Classifieds split.}
    \label{fig:task_analysis}
\end{figure}

We report performance on both benchmarks in Figure~\ref{fig:results_vwa} and Table~\ref{tab:results_wa}. We find:

\textbf{$\triangleright$ \methodname{} achieves state-of-the-art success rates.} \methodname{} attains the best average score (52.9\%), with large gains on Classifieds (64.1\%, +12.1 absolute over SGV) and Reddit (39.0\%, +6.0 absolute), while remaining competitive on Shopping (53.4\% vs.\ 57.0\% for SGV). Further, it nearly doubles the success rate of the Claude Computer Use baseline (which uses an image-based observation space), also outperforms strong baselines based test-time search and tool use by 15-20 points.

On WebArena, \methodname{} again achieves the highest overall average success rate on 5 of 6 splits (tied on the sixth), outperforming prior work in all domains by a large margin, and outperforming the best-performing skill-induction based method (ASI) by 9 points.

\textbf{$\triangleright$ Tools improve both success rates and efficiency.} In Figure~\ref{fig:results_vwa} (right), we demonstrate both the performance (measured by success rate) and efficiency (measured by average \# steps) of \methodname{} on each VisualWebArena split. As a control,  we benchmark our baseline implementation which uses an identical architecture but does not use tools. As seen, tools are crucial, improving performance by as much as 30.7\% (relative) and efficiency by 1.4x. The baseline agent's significantly lower success rates also validate that gains are not due to a stronger underlying LLM (GPT-5) alone.

\textbf{Performance ablations.} We ablate \methodname{} on VisualWebArena Classifieds (Table~\ref{tab:ablations}). We first vary the LLM execution agent, and find agents equipped with discovered tools are consistently more accurate and efficient (\emph{e.g.} GPT-5-mini: 7\% higher success rate, 27\% fewer steps). Stronger backbones benefit more, indicating that better reasoning improves tool selection and composition rather than low-level manipulation. Finally, we also benchmark a human demo strategy as a performance upper bound, wherein the authors manually demonstrate a set of tools rather than having the agent discover them - tools generated thus yield the highest success rate (66.0\%). Impressively, however, \methodname{} is able to recover most of this performance fully autonomously (64.1\%), with 5\% fewer steps.

Next, we ablate the two ancillary method components: we find that both multimodal DOM parsing (+2.6\%) and external verification (+3.3\%) yield modest performance gains, with the latter coming at the cost of extra checks (more steps). Combining all components yields the highest success (64.1\%), still with substantially fewer actions than baseline policies (21.3\% fewer steps).

\textbf{Analyzing ~\methodname{}.} In Figure~\ref{fig:task_analysis}, we perform a fine-grained analysis of our method on the Classifieds split. First, in Figure~\ref{fig:workflow_stats} we break down each discovered tool by the total step count of its action script and its distribution across step types and functionalities. We make the following observations: i) tools span a range of functionalities across communication, content management, and search, ii) tools with the shortest action scripts correspond to URL promotions (typically discover-oriented), whereas those with longer scripts skew heavily towards deterministic UI interactions (typically content-management \emph{e.g.} form-filling). iii) Agentic steps are rare: In fact, only 3 out of the 9 tools have at least one agentic step.

In Figure~\ref{fig:success_rate_analysis}, we analyze the success rates of each of these tools, measured by the ratio of successful tool invocations by the agent during the entire evaluation run. Tools are used frequently (\emph{e.g.} search listings is invoked 262 times) and achieve nearly perfect success rates, attesting to high reliability. Finally, Figure~\ref{fig:classifieds_action} breaks down the action type distribution of each tool for successful and failed agent trajectories -- as seen, the agent uses both primitive and tool actions extensively in both cases.

\begin{figure}[t]
    \centering
    \includegraphics[width=\textwidth]{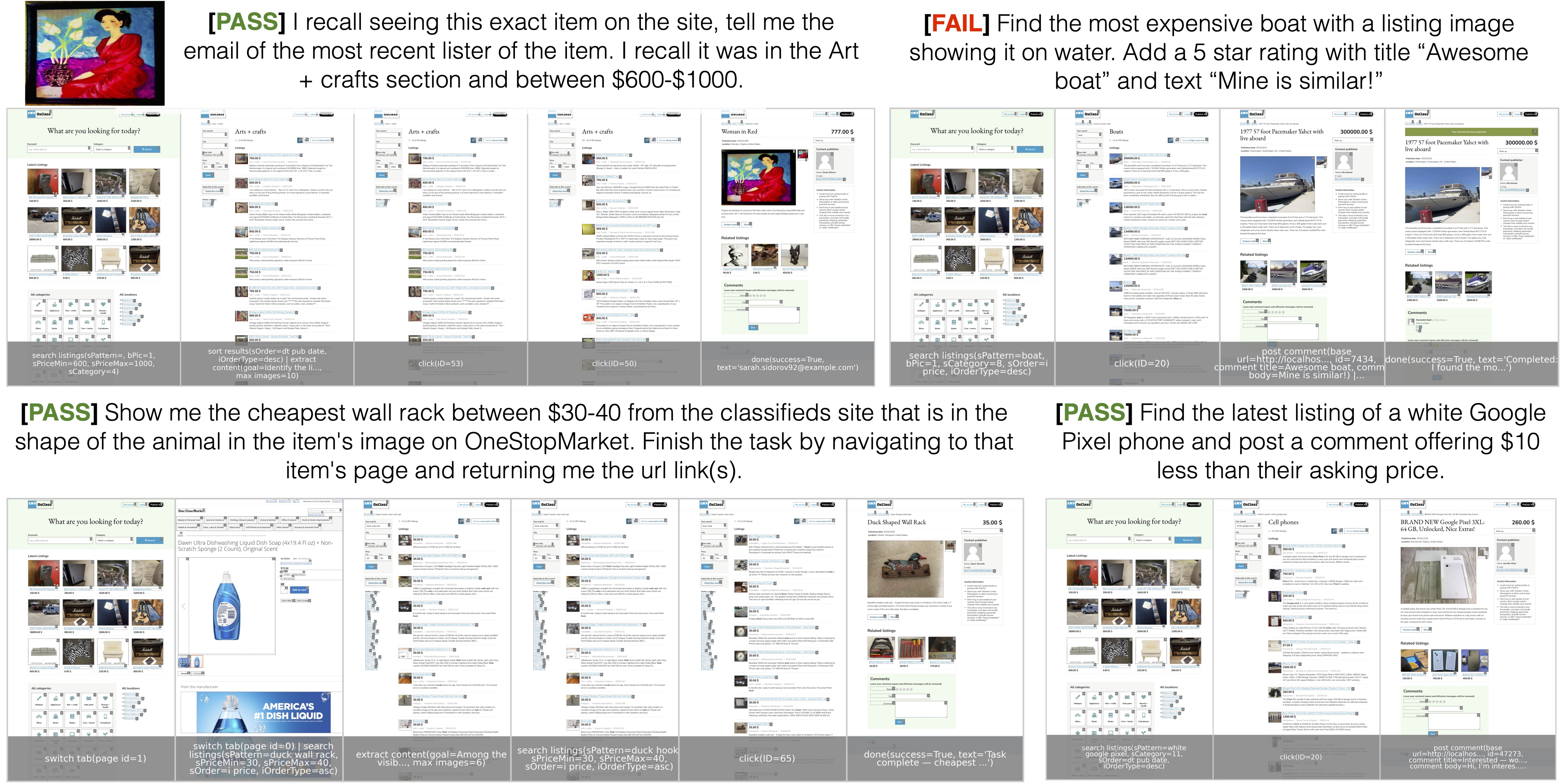}
    \caption{\textbf{Qualitative rollouts of \methodname{}.} Each row shows a task with tiled screenshots (left to right) and the agent's actions at each step (gray bars). \textbf{Left, top:} [PASS] ``Recall exact item and return the most recent lister's email.'' The agent chains \texttt{search\_listings}~$\to$~\texttt{sort\_results}~$\to$~\texttt{extract\_content}~$\to$~\texttt{click}, then surfaces the email from the item page. \textbf{Right, top:} [FAIL] ``Find the most expensive boat with an image showing it on water; rate 5 stars and comment.'' The agent finds expensive listings and grounds the the visual predicate but is unable to execute the rating action reliably. \textbf{Left, bottom:} [PASS] ``Cheapest wall rack between \$30--\$40 that matches the animal shape in the image.'' URL-level search and sorting prune the space, and an extraction step picks the correct visual match before navigation. \textbf{Right, bottom:} [PASS] ``Latest white Google Pixel; post a \$10-under offer.'' The agent locates the newest listing and uses \texttt{post\_comment} to complete the interaction. Across successes, trajectories are short (2--5 calls) and dominated by URL/navigation and schema-checked operations.}
    \label{fig:qualitative}
\end{figure}

Figure~\ref{fig:qualitative} presents  how \methodname{} composes a small set of learned tools to solve heterogeneous tasks. In successful cases, discovery tools (\texttt{search\_listings}, \texttt{sort\_results}) jump directly to the right result set via URL parameters, and an extraction step filters by visual or textual cues (e.g., ``animal-shaped'' wall rack) before a content action (e.g., \texttt{post\_comment}) finalizes the task. The traces show short programs (2--5 calls) with minimal UI clicking, explaining the step--count reductions reported in aggregate. The failure case exposes two common stressors: compound constraints that mix a global optimum (``most expensive'') with a fine-grained visual predicate (``image shows it on water''), and missing or site-gated side-effect tools (rating), which together exceed what the deterministic executor can guarantee without more agentic perception or broader tool coverage. Overall, the panel illustrates the intended division of labor---deterministic navigation and schema-checked operations for speed/robustness, with targeted agentic extraction when unavoidable.

\section{Discussion}
\label{sec:discussion}

In this work, we reframe browser automation around tools -- callable abstractions reverse-engineered from website functionality -- rather than agent-imagined skills implemented as a brittle sequence of UI actions. Our method ~\methodname{} exposes existing website functionality as robust tools that accept a validated input schema and accomplish a specific goal via a sequence of UI interaction, extraction, agent, and navigation steps, each with strong failsafes built in. ~\methodname{} achieves state-of-the-art performance on challenging web automation benchmarks while requiring fewer LLM interventions.

Our method has certain limitations. Offline tool discovery incurs an exploration and validation cost per-website, and the type and quality of the tools discovered is a function both of what our exploration uncovers and what the site exposes. Highly dynamic interfaces, A/B experiments, CAPTCHAs, and heavy anti-automation can reduce determinism or block URL promotion. Schemas may still miss rare parameter values; selector stabilization can drift after major redesigns; and some interactions (e.g., complex editors, file uploads) still require agentic steps. Our evaluation focuses on two research benchmarks, but broader external validity (\emph{e.g.}, enterprise apps) remains to be tested.

These limitations also present opportunities for future work. Online tool patching when selectors and schemas drift over time can improve robustness. Extracting canonical web patterns for common functionalities (e.g. search, filter, sort) can aid generalization. Hybrid integration with official APIs when available, external MCP servers~\citep{luo2025mcp}, and more agent-accessible observation spaces~\citep{lu2025build} can help further expand capabilities. Overall, our tool abstraction paradigm suggests a practical path for safe, auditable automation: tools carry explicit contracts, examples, and validation traces, making web agents easier to monitor, share, and maintain as sites evolve.

\textbf{Ethics Statement.} All authors have read and agree to the ICLR Code of Ethics. The benchmarks used (VisualWebArena and WebArena) are publicly available testbeds that simulate interactions with websites, and no experiments were conducted with human subjects. Our method is designed for research purposes; however, as with any browser automation technique, misuse (e.g., for scraping or spam) is possible. We emphasize that \methodname{} is intended to improve robustness and reproducibility of academic benchmarks, not to enable malicious automation. All data handling follows the licenses of the underlying benchmarks, and no private or user-sensitive data is involved.

\textbf{Reproducibility Statement.} We have made efforts to ensure reproducibility. The paper provides full details of the tool discovery and construction pipeline (Sec.\ref{sec:approach} and Sec.~\ref{sec:appendix}), optimization objectives and algorithmic design (Sec.\ref{sec:approach}), and benchmark setups (Sec.\ref{sec:experiments}). Implementation details, including model choices, observation formats, step limits, and verification procedures, are described in Sec.\ref{sec:experiments}. Appendix materials include pseudocode, algorithm tables, and ablation analyses. Our code will be made publicly available.

\bibliography{iclr2026_conference}
\bibliographystyle{plainnat}

\appendix
\section{Appendix}
\label{sec:appendix}

\subsection{Analysis}

\begin{figure}[t]
  \centering
  \includegraphics[width=\textwidth]{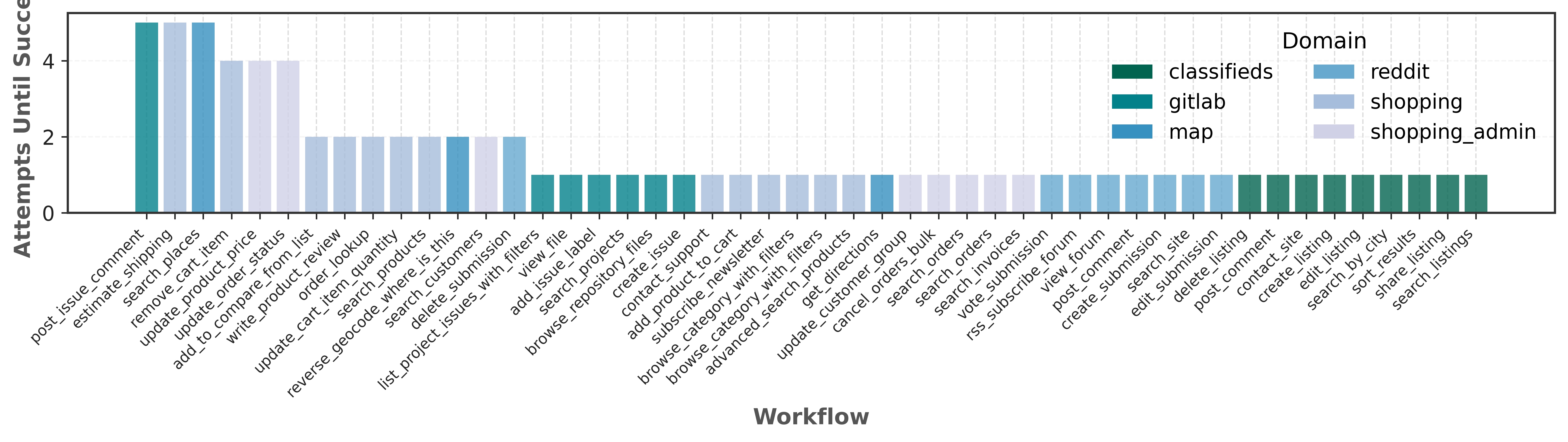}
  \caption{Number of tries until successful.}
  \label{fig:number_of_tries_until_success}
\end{figure}

\textbf{Tools.} In Figure~\ref{fig:number_of_tries_until_success}, we include a list of all tools discovered across the WebArena and VisualWebArena benchmarks, as well as the number of attempts required to obtain a validated implementation. As seen, most tools are discovered on the first attempt, but a few more nuanced functionalities (\emph{e.g.} post a comment on a Gitlab issue, searching on OpenStreetMaps, and estimating shopping on Shopping) require as many as 4 attempts.

\begin{figure}[t]
  \centering
  \begin{subfigure}[t]{0.55\textwidth}
    \centering
    \includegraphics[width=\textwidth]{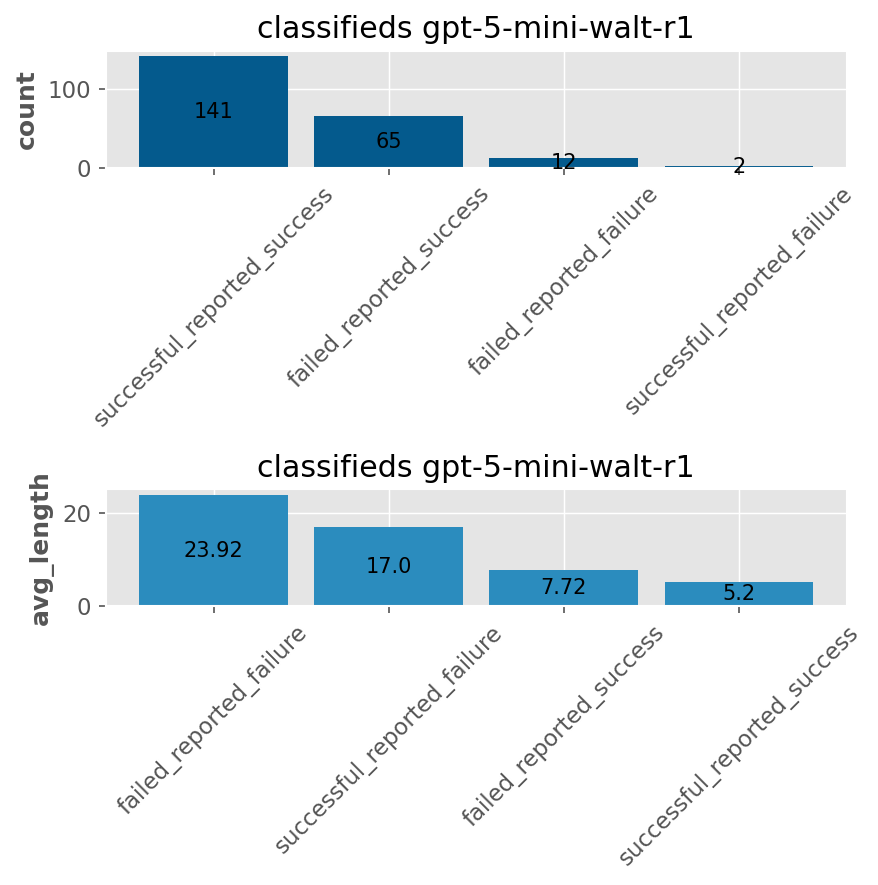}
    \caption{Self-reporting analysis}
    \label{fig:classifieds_self_reporting}
  \end{subfigure}
  \hfill
  \begin{subfigure}[t]{0.4\textwidth}
    \centering
    \includegraphics[width=\textwidth]{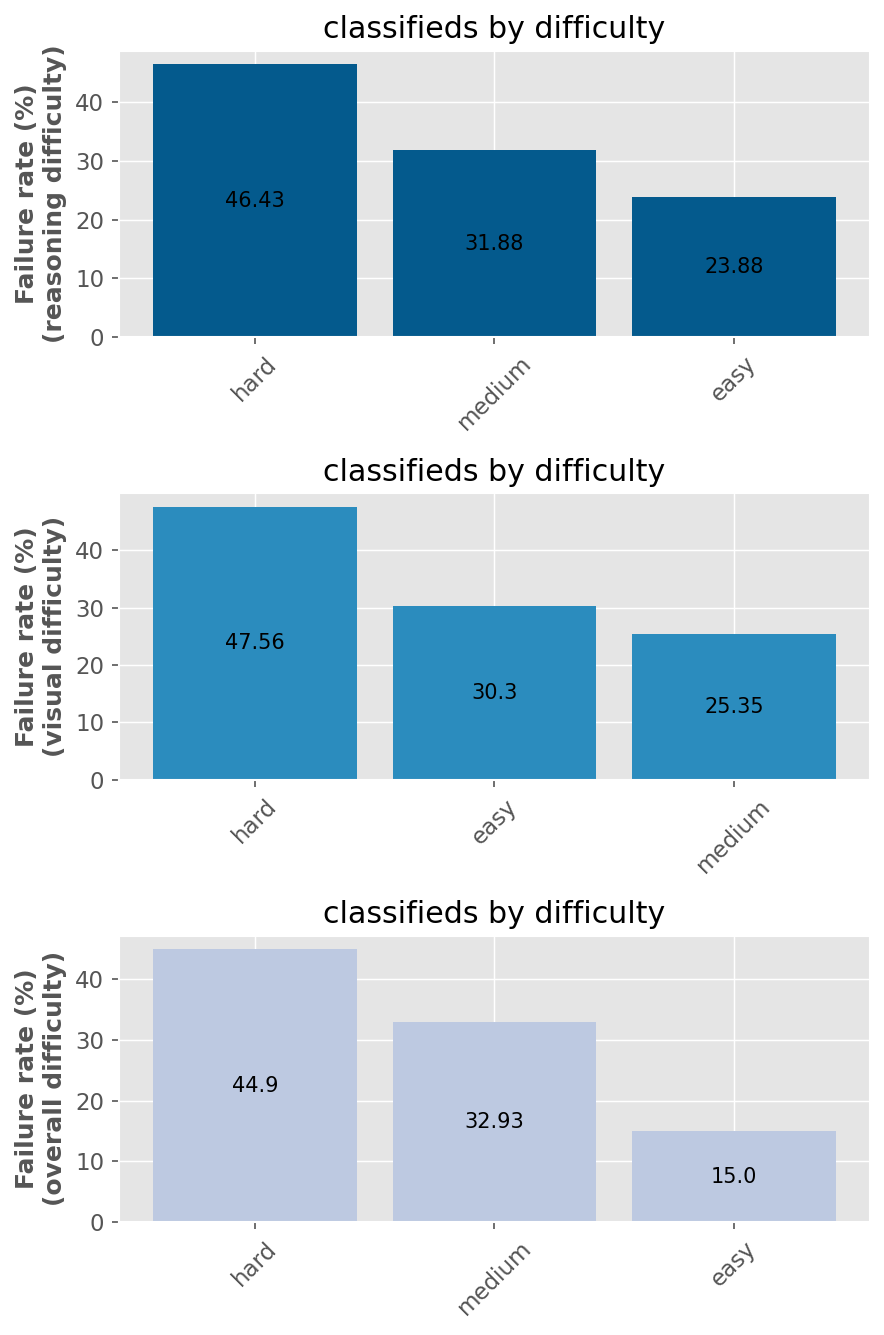}
    \caption{Difficulty analysis}
    \label{fig:classifieds_difficulty}
  \end{subfigure}
  
  \caption{Analysis of classifieds task performance across different dimensions.}
  \label{fig:classifieds_analysis}
\end{figure}

\textbf{Performance.} In Figure~\ref{fig:classifieds_self_reporting}, we include additional fine-grained performance analysis of our method on the Classifieds benchmark. First, we analyze the frequency and average length of successful and failed trajectories, segmented by the agent's own assessment of the task outcome -- as found in concurrent work~\citep{andrade2025let}, web agents suffer from an "agreement bias" and frequently rationalize even failed trajectories as successful. Our approach mitigates this bias by using an external verifier to corroborate the agent's assessment.

In Figure~\ref{fig:classifieds_difficulty}, we segment performance based on task difficulty (visual, reasoning, and overall), annotations for which are available in the benchmark.  Unsurprisingly, failure rates increase with increasing difficulty of any type - impressively though, \methodname{}'s failure rate does not cross 50\% even on the most difficult tasks.

\begin{figure}[t]
  \centering
  \includegraphics[width=\textwidth]{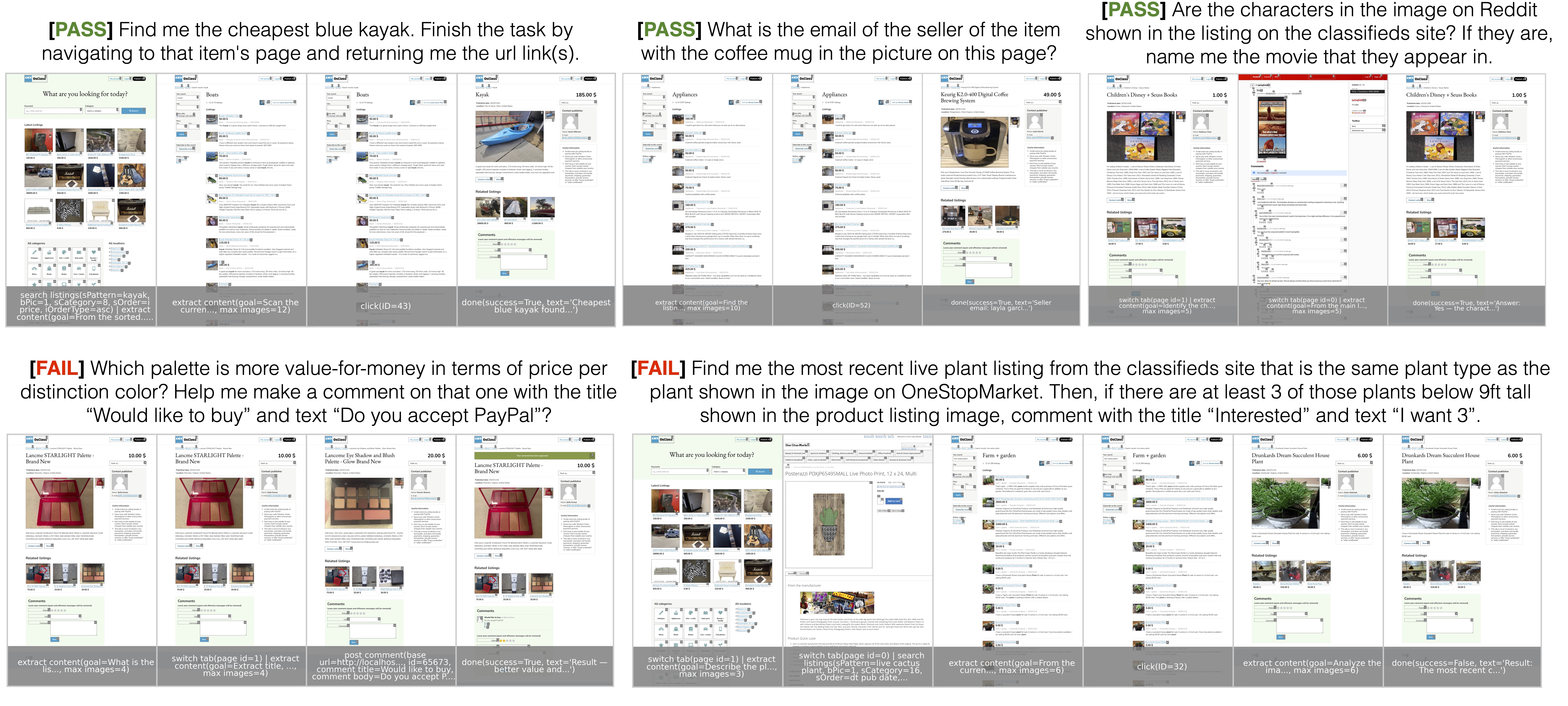}
  \caption{Qualitative examples showing \methodname{} tool discovery and execution on representative tasks.}
  \label{fig:qual_supp}
\end{figure}

\textbf{Qualitative Examples.} In Figure~\ref{fig:qual_supp}, we show additional qualitative examples of trajectory rollouts via our method, and include both successful and failure cases. Notably, our method is able to perform challenging visual grounding (\emph{e.g.} finding the coffee mug from a thumbnail image - top row, 2nd image) and reason across websites (top row, 3rd image). However, it still struggles with tasks requiring very complex multi-step reasoning (bottom row).


\begin{algorithm}[t]
  \caption{WALT: Two-Agent Tool Construction (Appendix)}
  \label{alg:appendix-walt}
  \begin{algorithmic}[1]
  \Require Candidate $\toolcandidate=(s_i,E_i,G_i)$, attempt budget $\maxattempts$
  \Ensure Validated tool $\toolopt\in\actionstools$ or $\text{FAIL}$
  
  \State $\text{attempts}\leftarrow 0$
  \While{$\text{attempts}<\maxattempts$}
    \State $\text{attempts}\leftarrow \text{attempts}+1$
  
    \State \textcolor{teal}{\textbf{Phase I: Exploration \& Stabilization (by $\agentbrowser$)}}
    \State $\traces \leftarrow \agentbrowser.\text{Execute}(s_i,E_i,G_i)$
    \If{$\traces=\text{FAIL}$} \State \textbf{continue} \Comment{retry with alternate exploration strategy} \EndIf
    \State $\traces \leftarrow \StabilizeSelectors(\traces)$ \Comment{resolve to stable DOM hashes/locators; drop unstable segments}
    \If{$\traces=\text{UNSTABLE}$} \State \textbf{continue} \EndIf
  
    \State \textcolor{teal}{\textbf{Phase II: Synthesis \& Optimization (by $\agenttool$)}}
    \State $\text{plan}\leftarrow\emptyset$
    \For{each segment $\xi \in \traces$}
      \State $\text{step} \leftarrow \agenttool.\text{ClassifyAndCreate}(\xi)$ \Comment{navigation / interaction / agentic}
      \State $\text{plan}\leftarrow \text{plan}\cup\{\text{step}\}$
    \EndFor
    \State $\text{plan}\leftarrow \AddAgenticFallbacks(\text{plan})$ \Comment{re-query DOM, retry alt selector, etc.}
    \State $\text{plan}\leftarrow \ReplaceWithURLOps(\text{plan})$ \Comment{promote eligible UI subsequences to URL ops}
    \State $\schema \leftarrow \InferSchema(\traces)$ \Comment{enums, optionals, descriptions, examples}
    \State $\testinputs \leftarrow \agenttool.\text{ExtractTestInputs}(\traces,\schema)$
    \State $\tool \leftarrow (\text{plan},\schema)$
  
    \State \textcolor{teal}{\textbf{Phase III: Registration \& Validation (by $\agentbrowser$)}}
    \State $\text{RegisterTool}(\tool,\schema,\testinputs)$
    \State $\text{result} \leftarrow \agentbrowser.\text{TestTool}(\tool,\testinputs)$
    \If{$\text{result}=\text{SUCCESS}$}
       \State \Return $\tool$ \Comment{validated; added to $\actionstools$ as $\toolopt$}
    \Else
       \State $\feedback \leftarrow \text{GetValidationErrors}(\text{result})$ \Comment{selector drift, missing enum, timeout, semantic mismatch}
       \State $(s_i,E_i,G_i) \leftarrow \RefineCandidate\big((s_i,E_i,G_i),\feedback\big)$ \Comment{update selectors, schema, or plan hints}
       \State \textbf{continue}
    \EndIf
  \EndWhile
  \State \Return \text{FAIL}
  \end{algorithmic}
  \end{algorithm}

\subsection{Implementation Details}
\label{subsec:implementation_appx}

\textbf{Tool Creation Agent Algorithm and System Prompt.} We include the system prompt for the tool discovery agent in Listing~\ref{fig:tool_exploration_agent_system_prompt} and algorithm and system prompt of the tool creation agent $\agenttool$ in Algorithm~\ref{alg:appendix-walt} and Listing~\ref{fig:tool_creation_agent_system_prompt}. 

\textbf{Baseline Implementation Details.} We use the Claude Computer-Use Agent with a dedicated desktop environment setup similar to OS-World~\citep{xie2024osworld}. Each task initializes with a Chrome browser and task-specific web pages. The agent receives desktop screenshots as observations, predicts OS-level actions, and executes them via pyautogui commands. Task completion is determined by either reaching the maximum step limit or agent prediction, with evaluation based on the final active webpage and parsed response.

We use \texttt{claude-4-sonnet-20250514} with thinking mode enabled (temperature=1). Due to Bedrock API limits, all screenshots and task images are resized to 1280$\times$720, with a maximum of 30 steps per task. Note that active webpage detection relies on heuristic algorithms using Playwright and Chrome DevTools Protocol, which may incorrectly identify the current page in edge cases. Reported accuracies should be viewed as lower bounds rather than exact measurements.

\subsection{Use of Large Language Models} 

Large language models (LLMs) were used to polish (proofreading, revising, and compressing) the writing, specifically Claude-4-Sonnet and GPT-5.

\newpage

\begin{tcolorbox}[colback=black!5!white, colframe=black!75, title=System Prompt of the Tool Discovery Agent, listing only, breakable]
  \label{fig:tool_exploration_agent_system_prompt}
  \centering
  \begin{lstlisting}[breaklines]
You are an expert browser automation agent designer. Your goal is to first systematically explore {base_url} and discover user-facing functionality offered by the website. Next, you will use this information to design a minimal but flexible API specification that captures these core user functions.

## Stage 1: Exploration

- Navigate systematically through user-facing site sections. For each area, ask: "What would a typical logged-in user want to accomplish here"?.
  - PRIORITIZE:
    - discovery & search (e.g. search, filters, categories, sorting)
    - content creation & management (e.g. create, edit, delete, view personal content)
    - communication & interaction (e.g. post comments, reply to comments, vote on content, share content)
    - organization (e.g. save favorites, manage lists, subscribe to alerts)

Exploration Guidelines:
- You are already logged in with full user access to the site.
- Only document tools that actually exist and function on the site.
- Aim to explore atleast 10-20 **diverse** tools covering comprehensive user functionality

## Stage 2: API Design
- In this stage, you will use the information from the exploration stage to design a minimal but diverse and flexible API **specification** that captures these core user functions.
- **API Design principles**:
  - **Goal-oriented**: Focus on user goals, not UI mechanics. One clear goal per function. Good candidates typically compose an active verb and noun (eg. create+listing, post+comment, search+forums, etc.) 
  - **Reusable**: Functions should be parameterizable and work with ANY item/content, not hardcoded specifics
  - **Composable**: Propose modules with **diverse** functionality that can be **combined** to achieve more complex goals

        
API Design Guidelines:
- Use the information gathered from the exploration stage extensively
- DO NOT TRY TO EXPLORE THE SITE AGAIN IN THIS PHASE.
- Do not worry about implementation details, as long as you have confirmed the underlying functionality exists.

FINAL OUTPUT FORMAT: Return a **single valid JSON object** with the following fields for each proposed function:
1. **name**: Strategic goal identifier (e.g. "edit_listing", "search_by_category")
2. **start_url**: Exact URL where tools begins (only URLs you've actually visited)
3. **description**: Goal with parameterization (e.g. "locate listing by user-provided title and update its properties to user-provided values")
4. **elements**: Key interactions (type and purpose, with available options for dropdowns/menus - does not need to be exhaustive or perfect)

{{
  "tools": [
    {{
      "name": "strategic_tools_name", 
      "start_url": "https://example.com/some/page",
      "description": "Accomplish specific goal with user-provided parameters",
      "elements": [
        {{"type": "input", "purpose": "enter user-provided search terms"}},
        {{"type": "select", "purpose": "choose user-specified category", "options": ["Electronics", "Clothing", "Books", "All Categories"]}},
        {{"type": "select", "purpose": "sort results", "options": ["Newly listed", "Lower price first", "Higher price first"]}},
        {{"type": "button", "purpose": "submit search"}}
      ]
    }}
  ]
}}
  \end{lstlisting}
  \end{tcolorbox}
  \begin{tcolorbox}[colback=black!5!white, colframe=black!75, title=System Prompt of the Tool Creation Agent, listing only, breakable]
  \label{fig:tool_creation_agent_system_prompt}
  \centering
  \begin{lstlisting}[breaklines]
You are a master at building re-executable tools from browser automation steps. Your task is to convert a sequence of Browser Use agent steps into a parameterized reusable tool.

**Core Objective**
Transform recorded browser interactions into a structured tool by:
- Extracting actual values (not placeholder defaults) from the input steps
- Identifying reusable parameters that should become tool inputs
- Creating deterministic steps wherever possible
- Optimizing the tool for clarity and efficiency
- Optimize Navigation: Skip unnecessary clicks when direct URL navigation works

**Input Format**
You will receive a series of messages, each containing a step from the Browser Use agent execution:

**Step Structure**
Each message contains two parts:
- parsed_step (content[0]) - The core step data:    
  - url: Current page URL
  - title: Page title
  - agent_brain: Agent's internal reasoning
    - evaluation_previous_goal: Success/failure assessment of previous action
    - memory: What's been accomplished and what to remember
    - next_goal: Immediate objective for next action
  - actions: List of actions taken (e.g., go_to_url, input_text, click_element, extract_content)
  - results: Outcomes of executed actions with success status and extracted content
  - interacted_elements: DOM elements the agent interacted with, including selectors and positioning
      - special field element_hash: unique identifier for elements the agent interacted with.
- screenshot (content[1]) - Optional visual context of the webpage
---------------------------------------------------------------------------------------------------
**Output Requirements**

1. Tool Analysis (CRITICAL FIRST STEP)
The tool_analysis field must be completed first and contain:
- Step Analysis: What the recorded steps accomplish overall
- Task Definition: Clear purpose of the tool being created
- Action Plan: Detailed to-do list of all necessary tool steps
- Variable Identification: All input parameters needed based on the steps and task
- Step Optimization: Review if steps can be combined, simplified, or if any are missing. Always prefer: 1) Navigation steps (where possible), 2) Deterministic steps (when elementHash is stable), 3) Agent steps only as last resort for truly dynamic content.

**Input Schema:** Define tool parameters using simple JSON schema
- Include at least one input unless the tool is completely static
- Add descriptive documentation: Always include desriptive field explanations
- **Field Requirements (setting "required" true/false):** Match website requirements - if website requires it, tool requires it

**Steps Array**    
Each step must include a "type" field and a brief "description".

** Tool DESIGN PRINCIPLES:**
- Sequential & Deterministic: Steps execute in order, no conditional branching
- Single Purpose: Each tool accomplishes ONE specific task
- No Optional Logic: Avoid "if user wants X, then do Y" patterns
- Essential Steps Only: Every step must be required for the core task
- Parameter-Driven: Use input parameters to customize behavior, not conditional steps
---------------------------------------------------------------------------------------------------
**Step Creation Algorithm (Two-Pass Approach)**    
This tool generation uses a two-pass approach: PASS 1 creates basic steps using simple rules, then PASS 2 (optional) potentially optimizes it by replacing UI interaction sequences with more efficient URL manipulation, if possible.

**PASS 1: Basic Step Generation (Rule-Based):** Follow this exact sequence for each agent action - no decisions required:

### STEP 1: Classify Action Type

FOR each agent action:
  IF navigation/URL changes then Navigation Algorithm
  ELIF extracts data then Extraction Algorithm
  ELIF UI interaction:
    IF elementHash exists then Deterministic Interaction
    ELSE IF essential then Agentic Interaction
    ELSE then Skip
  ELSE then Skip
STEP 2: Execute the Appropriate Algorithm

**Navigation Algorithm:** Creates navigation steps to move between pages or change URLs
- url: Target URL to navigate to
- description: Brief explanation of the navigation purpose

**Extraction Algorithm:** Extracts goal-relevant data or content from the current page
- goal: Description of what data to extract from the page
- output: Label for the captured data (use meaningful names like "listing_data", "search_results")
- description: Brief explanation of what data is being extracted

**Deterministic Interaction Algorithm:** Interacts with page elements using stable identifiers
- elementHash: Unique identifier for the DOM element (required - stable selectors auto-generated)
- value: Text to input (for input steps)
- selectedText: Option to select (for select_change steps)
- key: Key to press (for key_press steps, e.g., 'Tab', 'Enter')
- scrollX, scrollY: Pixel offsets for scrolling (for scroll steps)
- description: Brief explanation of the interaction purpose
- seconds: Number of seconds to sleep (for wait steps)

**Agentic Interaction Algorithm:** Handles dynamic interactions requiring reasoning
- task: User perspective goal (e.g., "Select restaurant named {{{{restaurant_name}}}}")
- description: Why agentic reasoning is needed and what the step accomplishes
- max_steps: Always specify limit (3-8 typical, never null)

**[Optional] PASS 2: URL Manipulation Optimization**
REPLACE UI interaction sequences in tool with a single URL navigation for better efficiency and reliability
- Web functionalities (typically GET requests eg. search, filtering, sort, pagination) are often achievable by navigating to URL modified with certain parameters
- By inferring these parameters correctly, tools requiring several UI interactions can be accomplished in only a few steps
---------------------------------------------------------------------------------------------------

**Context:**
Task Goal: {goal}
Available Actions: {actions}

The goal shows the original task given to the agent. Assume all agent actions can be parameterized and identify which variables should be extracted. Input session events will follow in subsequent messages.
  \end{lstlisting}
  \end{tcolorbox}

\end{document}